\documentclass[10pt,twocolumn,letterpaper]{article}

\usepackage[pagenumbers]{iccv} 

\usepackage{adjustbox} 
\usepackage{tabularx} 
\usepackage{orcidlink}
\usepackage[table]{xcolor} 
\usepackage{booktabs}
\usepackage{multirow}
\usepackage{float}

\definecolor{iccvblue}{rgb}{0.21,0.49,0.74}

\def\paperID{24} 
\def\confName{ICCV}
\def\confYear{2025}

\title{	PU-Gaussian: Point Cloud Upsampling using 3D Gaussian Representation}

\author{
Mahmoud Khater\textsuperscript{1,2} \orcidlink{0009-0008-6355-9510}
\quad Mona Strauss\textsuperscript{1,2} \orcidlink{0009-0000-3007-6709}
\quad Philipp von Olshausen\textsuperscript{2}
\quad Alexander Reiterer\textsuperscript{1,2}
\vspace{5pt}
\\
\textsuperscript{1}University of Freiburg, Germany
\textsuperscript{2}Fraunhofer IPM, Freiburg, Germany
}

\begin{document}
\maketitle
\begin{abstract}
Point clouds produced by 3D sensors are often sparse and noisy, posing challenges for tasks requiring dense and high-fidelity 3D representations. Prior work has explored both implicit feature-based upsampling and distance-function learning to address this, but often at the expense of geometric interpretability or robustness to input sparsity.
To overcome these limitations, we propose PU-Gaussian, a novel upsampling network that models the local neighborhood around each point using anisotropic 3D Gaussian distributions. These Gaussians capture the underlying geometric structure, allowing us to perform upsampling explicitly in the local geometric domain by direct point sampling. The sampling process generates a dense, but coarse, point cloud. A subsequent refinement network adjusts the coarse output to produce a more uniform distribution and sharper edges. We perform extensive testing on the PU1K and PUGAN datasets, demonstrating that PU-Gaussian achieves state-of-the-art performance. We make code and model weights publicly available at \url{https://github.com/mvg-inatech/PU-Gaussian.git}.

\end{abstract}


\section{Introduction}
\label{sec:intro}
Point clouds are a fundamental 3D data representation, simple, efficient, and directly captured by sensors such as LiDAR and RGB-D cameras \cite{guo2020deep}. Unlike structured formats such as meshes or voxels, point clouds remain lightweight and versatile, powering applications from autonomous driving and robotics to augmented reality and object recognition. However, raw point clouds often suffer from issues like low resolution, non-uniform distribution, and noise, due to hardware limitations and various environmental factors \cite{yu2018pu}.

\begin{figure}[t]
\centering
\includegraphics[width=\linewidth]{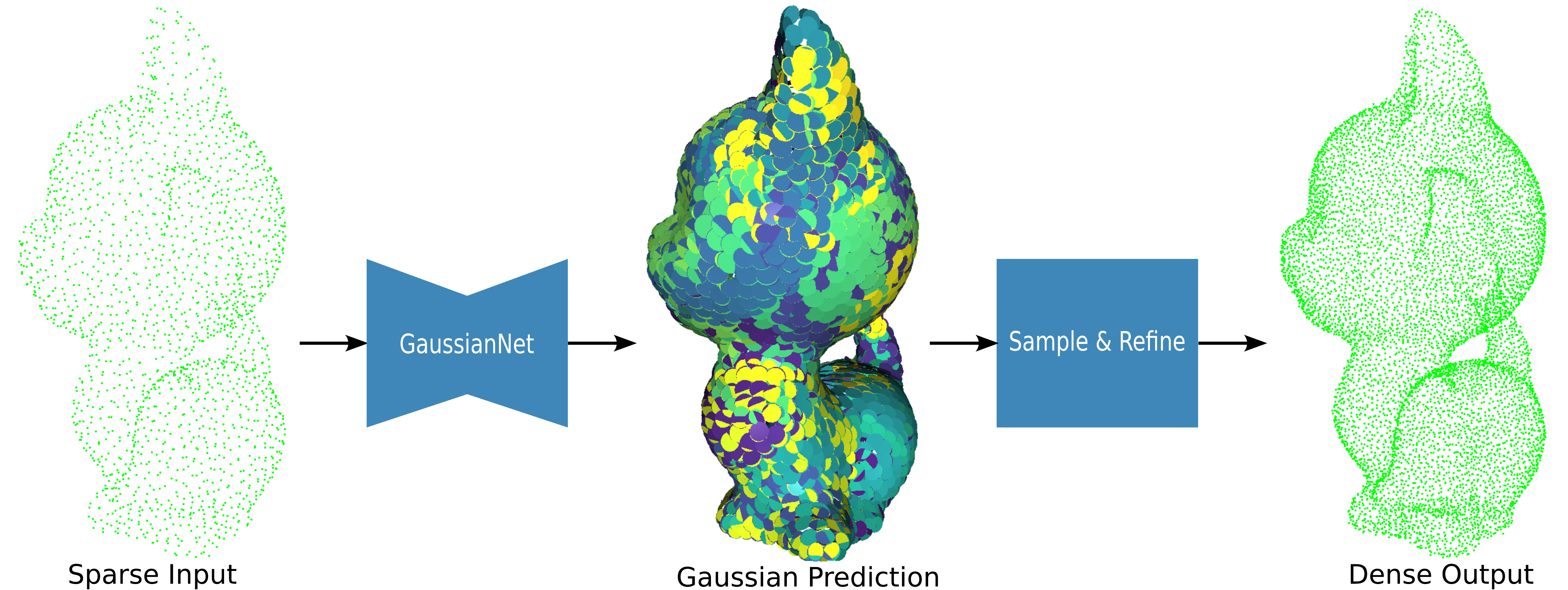}
\caption{PU-Gaussian: a novel point-cloud upsampling pipeline. First, our Gaussian network predicts an anisotropic 3D Gaussian for each local patch in the sparse input. Next, we sample points from these Gaussians and refine their positions using a lightweight refinement network. Compared to prior methods, PU-Gaussian offers a more direct and interpretable approach.}
\label{fig:teaser}
\end{figure}

High-quality, dense, and uniformly distributed point clouds are crucial for downstream tasks, such as 3D semantic segmentation, completion, and surface reconstruction. Sparse or irregular point sets can significantly hinder performance, especially when fine-grained geometric details are necessary for accurate interpretation. Consequently, developing effective point cloud upsampling methods to transform sparse point clouds into denser representations has become a crucial area of research in 3D deep learning.

Following the pioneering work of PU-Net \cite{yu2018pu}, numerous deep learning methods for point cloud upsampling have emerged, which can be categorized into two main approaches: direct prediction of point clouds through feature learning and geometric reconstruction. Early methods relied on feature space upsampling techniques, such as node shuffling, folding, and attention mechanisms \cite{qian2021pu,yu2018pu,li2021point,luo2021pu,yifan2019patch,li2019pu,li2021point,qiu2021pu,long2022pc2}; however, they typically required training fixed models for different upsampling ratios. Additionally, upsampling in feature space lacks interpretability and can lead to point condensation near the original sparse cloud. The second approach involves generating a coarse point cloud and then iteratively refining it through the use of a learned local distance indicator \cite{He_2023_CVPR, li2024APU-LDI}. However, this method can cause computational overhead through the iterative process. It is inherently more challenging due to the difficulties in learning a distance metric from a sparse input \cite{du2024arbitrary}.

To address the limitations, we propose PU-Gaussian, a geometry-aware approach based on explicit local surface modeling using anisotropic Gaussians. Drawing inspiration from Gaussian splatting techniques \cite{kerbl3Dgaussians}. PU-Gaussian fits a local 3D anisotropic Gaussian distribution around each point in the input cloud, enabling direct sampling in 3D space and capturing fine-grained surface geometry with high fidelity.

In our approach, we formulate the problem as one of local distribution fitting by learning to predict Gaussian primitives around each point in the sparse point cloud. Specifically, for each input point \(x_i\), we estimate an anisotropic Gaussian distribution \(\mathcal{N}(\mu_i, \Sigma_i)\), which approximates the underlying local surface geometry from which \(x_i\) was sampled. We then draw \(r\) samples from each of these Gaussians, yielding a coarse, denser point cloud. Our method consists of a two-stage architecture. In the first stage, the Gaussian network estimates the Gaussian parameters for local surface patches and generates a coarse upsampled point cloud by sampling from these distributions. In the second stage, a refinement network adjusts the sampled point positions to preserve structural details and improve reconstruction accuracy.
We compare PU-Gaussian against a set of state-of-the-art deep-learning baselines for point cloud upsampling on PUGAN \cite{li2019pu} and PU1K \cite{qian2021pu} datasets.
Concretely, we make the following contributions:
\begin{enumerate}[topsep=0pt, partopsep=0pt, parsep=0pt, itemsep=0pt]
  \item We introduce PU-Gaussian, a novel neural network approach for upsampling point clouds using Gaussian representation.
  \item Evaluate against state-of-the-art deep-learning baselines on the PUGAN and PU1K benchmarks.
  \item Release code, data, and model weights at \url{https://github.com/mvg-inatech/PU-Gaussian.git}.
\end{enumerate}
\section{Related work}
One of the earliest and most influential methods is PU-Net  \cite{yu2018pu}, which leverages the PointNet++ \cite{qi2017pointnet++} architecture for feature extraction. PU-Net established a three-stage framework—feature extraction, feature expansion, and geometric reconstruction—that has become a cornerstone in the field. Subsequent methods have built on this foundation: for example, PU-GAN  \cite{li2019pu} incorporates Generative Adversarial Networks to promote a more uniform point distribution, while PUFA-GAN \cite{liu2022pufa} enhances edge detail capture through a frequency-aware discriminator. PUGCN \cite{qian2021pu} departs from traditional folding techniques and introduces node shuffling for feature expansion. Similarly, PU-EVA \cite{luo2021pu}, integrates attention-based upsampling to support arbitrary scaling. Additionally, PU-Transformer \cite{qiu2021pu} explores multistage Transformer-based feature extraction, and PU-EdgeFormer \cite{kim2023pu} introduces an edgeconv unit combined with an attention mechanism.

Several approaches—including PU-CRN, Dis-PU, and MPU \cite{du2022point,li2021point, yifan2019patch}—follow the three-stage framework but incorporate multi-stage upsampling strategies. MPU, for instance, uses a progressive upsampling approach with a series of sub-networks that refine different levels of detail, gradually improving the coarse point cloud. PU-CRN employs a cascaded refinement network where multiple stages of feature extraction, expansion, and coordinate reconstruction work iteratively to enhance point cloud density and quality.

\noindent\textbf{Distance metric learning}: Another common approach is to learn a distance metric that represents the distance from a query point to the dense point cloud. Grad-pu \cite{He_2023_CVPR} learns a point-to-point network and refines the result using Newton step optimization. APU-LDI \cite{li2024APU-LDI}, learns a Signed distance Function for each surface and project query points on the surface using a distance network.

\noindent\textbf{Local patch representation}: PU-VoxelNet \cite{du2024arbitrary} voxelizes the input and learns a density distribution for each voxel. The main difference with our approach is that the voxels are discrete and rigid compared to a Gaussian representation that is free to move contentiously. RepKPU \cite{rong2024repkpu} learns a local kernel representation and uses a kernel-to-displacement module to generate the upsampled point cloud. APU-SMG  \cite{dell2022arbitrary} maps the input into a spherical mixture of Gaussian distributions and uses a decoder to reconstruct the data back into 3D space. In contrast, we fit our Gaussians in the 3D space which allows us to sample directly without any decoding stage.

In our approach, we offer an interpretable expansion stage by learning a local Gaussian distribution that represents the neighborhood of each point, we then sample points from this distribution and pass them to a refinement network that enhance spatial accuracy and geometric consistency of the sampled points. This design aims to enhance uniformity, as we can control the sampling process and preserve fine geometric details more effectively.

\section{Methodology}
In this section, we begin with a formal problem definition and outline the perspective that motivated our proposed solution. We then provide a detailed description of the pipeline in the subsequent subsections.

\subsection{Problem definition}




 Given a sparse point cloud \(\mathcal{X} = \{x_i\}_{i=1}^N\), where each point \(x_i \in \mathbb{R}^3\) is assumed to be stochastically sampled from an underlying 3D surface \(\mathcal{S}\). Our primary objective is to generate a dense uniform set of points \( Q = \{q_j\}_{j=1}^{rN} \) such that \( Q \) represent the same underlying surface \( S \), where \( r \in \mathbb{Z} \) and \( r > 1 \).
We view each point \(x_i\) in the sparse point cloud as the result of a single draw from a local distribution \(p_i(x)\) centered around a point on \(\mathcal{S}\). Our goal is to learn this distribution \(p_i\), and then sample \(r\) points from it to densify the point cloud. To ensure geometric consistency and interpretability, we require that each \(p_i\) be a \emph{direct} probability distribution in \(\mathbb{R}^3\), eleminating the need for mappings from latent-dimensional spaces. This allows for straightforward sampling and a meaningful spatial interpretation of the generated points. We choose to parameterize this local distribution as an anisotropic Gaussian for its ability to produce a geometrical consistent view, as evident in the gaussian splatting paper \cite{kerbl3Dgaussians}.

\begin{figure*}[t]
    \begin{minipage}[t]{\linewidth}
        \centering
        \includegraphics[width=\linewidth]{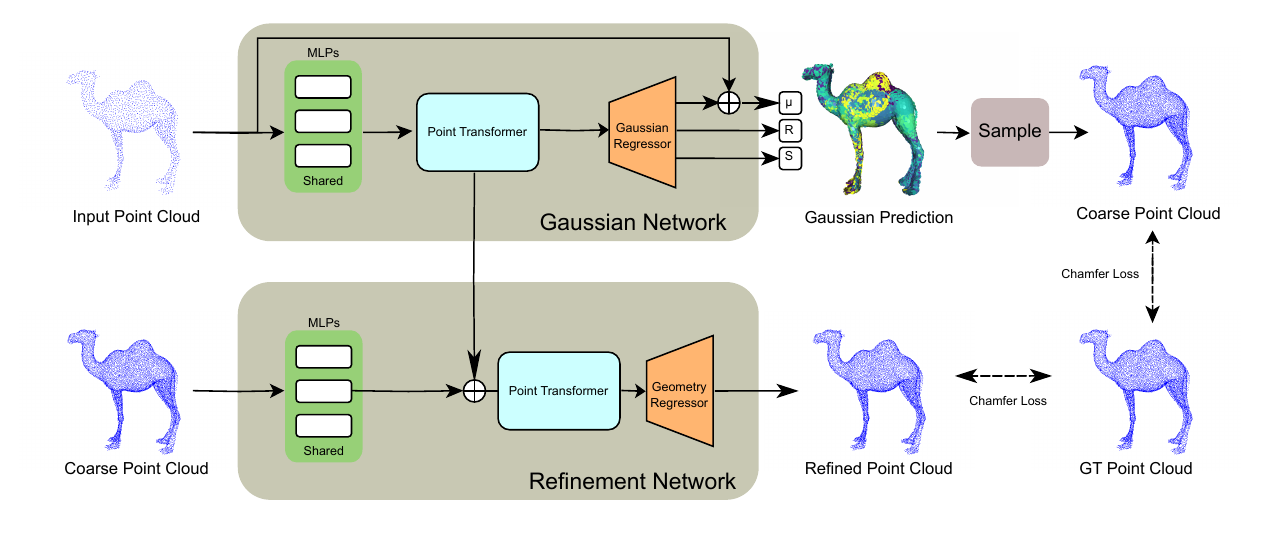}
        \vspace{1mm}
        \caption{Overview of our pipeline, We first pass the sparse Input point cloud to our gaussian prediction network. We then sample from the gaussian prediction a coarse point cloud. We pass the point cloud to our refinement network to better preserve the edges.}
        \label{fig:pipeline}
    \end{minipage}
\end{figure*}

\subsection{Preliminaries}

3D Gaussians are highly effective in capturing local geometric shapes within a three-dimensional space. Their flexibility comes from the ability to stretch along each axis and rotate the orientation of these axes, allowing them to adapt to various shapes. This characteristic has been used in \cite{kerbl3Dgaussians}, which utilizes Gaussians to represent local primitives in scene reconstruction. An anisotropic Gaussian is parametrized by a mean vector $\mu$ and a covariance matrix $\Sigma$. The covariance matrix can be decomposed into two factors: a rotation matrix and a scale matrix, which can be optimized independently allowing for an easier, more stable optimization. The covariance matrix $\Sigma$ must be positive semi-definite and is given by:

\begin{equation}
\Sigma = R S S^\top R^\top
\label{eq:covariance}
\end{equation}

where $S \in \mathbb{R}^{3 \times 3}$ diagonal matrix that defines the spread along the $x$, $y$, and $z$ directions, and $R  \in \mathbb{R}^{3 \times 3}$ rotation matrix that defines the Gaussian Axes orientation in 3D space. Constructing the covariance matrix in this way guarantees that it is always positive semidefinite during optimization. The anisotropic Gaussian function in 3D is defined as:

\begin{equation}
G(\mathbf{x}) = \alpha \exp\left( -\frac{1}{2} (\mathbf{x} - \mu)^\top \Sigma^{-1} (\mathbf{x} - \mu) \right)
\label{eq:gaussian_function}
\end{equation}

where $\mathbf{x}$ is a point in 3D space, $\mu$ is the mean vector, $\alpha$ is a scalar normalization factor and $\Sigma$ is the covariance matrix.

\subsection{Pipeline Overview}




Our approach consists of three primary stages: (1) a Gaussian prediction network that estimates the parameters of the gaussian \(\mu_i\) and \(\Sigma_i\) for each input point. (2) a sampling module that draws \(r\) points from each learned gaussian distribution yielding a dense but coarse point cloud, and (3) a refinement network that improves the quality and accuracy of the generated dense point cloud.

\subsection{Gaussian Network}

The Gaussian network is responsible for predicting the three parameters that define the gaussian for each point in space: scale, rotation, and mean.
The input point cloud $P_{\text{in}} \in \mathbb{R}^{N \times 3}$ is passed through a projection Multi-Layer Perceptron (MLP) that maps the input from 3D into a feature space of dimension $C$. The projected points are then processed by Point Transformer \cite{zhao2021point}, generating a feature vector $F \in \mathbb{R}^{N \times C}$. This feature vector serves as a local and global descriptor for each point. We employ Point Transformer as our backbone feature extractor due to its lightweight architecture and its ability to model local geometric structures without relying heavily on computationally expensive K-nearest neighbor operations. 

\begin{equation}
F = \text{PointTransformer}(\text{MLP}_{\text{proj}}(P_{\text{in}}))
\label{eq:feature_extraction}
\end{equation}

The extracted feature vector $F$ is then passed to our Gaussian Regressor Network. The network consists of three separate heads \( H_s \), \( H_r \), and \( H_{\text{offset}} \) that predict the parameters that define a Gaussian for each point:

\begin{equation}
\begin{split}
S = \text{softmax}(H_s(F)), \quad R = H_r(F), \\
\Delta = H_{\text{offset}}(F), \quad \mu = \Delta + P_{\text{in}}.
\end{split}
\label{eq:gaussian_params}
\end{equation}





Where $S \in \mathbb{R}^{N \times 3}$ represents the diagonal entries of a per-point scale matrix. $R \in \mathbb{R}^{N \times 4}$ represents the quaternion encoding of the rotation matrix. $\Delta \in \mathbb{R}^{N \times 3}$ represents the offset that translates the input point $P_{\text{in}}$ to a refined mean position $\mu$ for the Gaussian. The three heads, \( H_s \), \( H_r \), and \( H_{\text{offset}} \), each consist of two 1D convolutional layers with a kernel size of 1, separated by a Leaky ReLU activation function. The output of \( H_s \) is further processed by a softmax function to ensure the resulting scale parameters form a normalized distribution.

\subsection{Sampling}

After predicting a Gaussian distribution for each point, we sample \( r \) points from each Gaussian. To enable gradient-based optimization and end-to-end training, we employ the reparameterization trick \cite{kingma2015variational} during the sampling process. Let \( \mu \) denote the mean, \( \Sigma \) the covariance matrix as defined in equation~\ref{eq:covariance}, and \( \epsilon \) a random variable drawn from a standard normal distribution. The sampled points are then computed as:

\begin{equation}
P_{\text{coarse}} = \mu + \Sigma \odot \epsilon
\label{eq:sampling}
\end{equation}

where \( \mu \in \mathbb{R}^{N \times 3} \) is the predicted mean, \( \Sigma \in \mathbb{R}^{N \times 3 \times 3} \) is the covariance matrix, and \( \epsilon \in \mathbb{R}^{N \times r \times 3} \) is a noise tensor sampled from a standard normal distribution. To enforce tighter control over the shape of the Gaussians, any samples exceeding two standard deviation from the mean are discarded. As a result, the final set of sampled points has shape \( P_{\text{coarse}} \in \mathbb{R}^{N \times r \times 3} \), where \( N \) denotes the number of input points, and \( r \) is the number of samples drawn per Gaussian.

\subsection{Refinement Network}

Similar to prior work~\cite{du2022point, li2021point}, the goal of our refinement network is to enhance the positional accuracy of the coarse point cloud prediction while preserving important geometric structures such as edges and surfaces. Our refinement module is based on the architecture proposed by~\cite{du2022point}.

The refinement process operates on the coarse point cloud \( P_{\text{coarse}} \) and the feature descriptors \( F \), which are extracted from the initial Point Transformer within our Gaussian Network. We begin by projecting \( P_{\text{coarse}} \) into the same latent space \( C \) as \( F \) using a multi-layer perceptron: \( \text{MLP}_{\text{proj}}(P_{\text{coarse}}) \). To ensure alignment between features and sampled points, \( F \) is replicated \( r \) times using an interleaving operation. The projected coordinates and repeated features are then concatenated and passed through a Point Transformer to produce refined features \( F' \in \mathbb{R}^{rN \times C} \): 
\begin{equation}
F' = \text{PointTransformer}(F \oplus \text{MLP}_{\text{proj}}(P_{\text{coarse}}))
\label{eq:feature_extraction_2}
\end{equation}

The resulting feature representation \( F' \) is passed through an offset regressor, which predicts a residual offset \( \Delta \) for each point. These offsets are added to the initial coarse points to obtain the final refined point cloud: \( P_{\text{up}} = P_{\text{coarse}} + \Delta \). During training, we make predictions using the output of a single refinement stage. However, at inference time, we pass the output through the refinement stage one more time. This additional step is possible because the model’s inherent stochasticity during sampling allows it to be robust while
passing the output for one more refinement stage.

\subsection{Loss Formulation}

We adopt a two-stage loss formulation to supervise both coarse point cloud generation
and refined point cloud prediction. In the first stage, we evaluated the accuracy of the
prediction using the Chamfer distance as a reconstruction metric. To regularize, the Gaussian
we try to enforce a match between the mean and $r$ points in the ground truth. Specifically,
for each ground truth point, we identify its nearest predicted Gaussian mean and compute
the following metric:
\begin{equation}
L_{\text{Gaussian}} = \frac{1}{N} \sum_{i=1}^{N} (\mathbf{x}_i - \boldsymbol{\mu}_i)^\top \Sigma_i (\mathbf{x}_i - \boldsymbol{\mu}_i)
\label{eq:mahalanobis_loss}
\end{equation}

where \( N \) is the total number of points in the ground truth point cloud, \( \mathbf{x}_i \) denotes the \( i \)-th ground truth point, and \( \boldsymbol{\mu}_i \) is its corresponding predicted Gaussian mean.

In the second stage, we only apply chamfer Loss on the the refined output \( P_{\text{up}} \) with respect to the ground truth point cloud \( P_{\text{gt}} \).



\subsection{Training and Inference}
Our network undergoes a two-stage training process. Initially, we train the Gaussian
network independently for 100 epochs, optimizing with Chamfer loss and a regularization
loss. Following this, the complete network is trained end-to-end for another 100 epochs
across both the PUGAN and PU1K datasets. We train our model with a $6\times$ upsampling factor and subsequently downsample the output to $4\times$ using Farthest Point Sampling. This strategy yields improved upsampling performance. During inference, the refinement stage is executed twice. The first pass uses the coarse output from the Gaussian network, while the
second pass utilizes the output generated by the preceding refinement step. We found that
this approach gives a more stable output due to the stochasticity of the sampling process.
\begin{figure*}[t]
\centering
\includegraphics[width=0.85\textwidth,height=0.7\textheight,keepaspectratio]{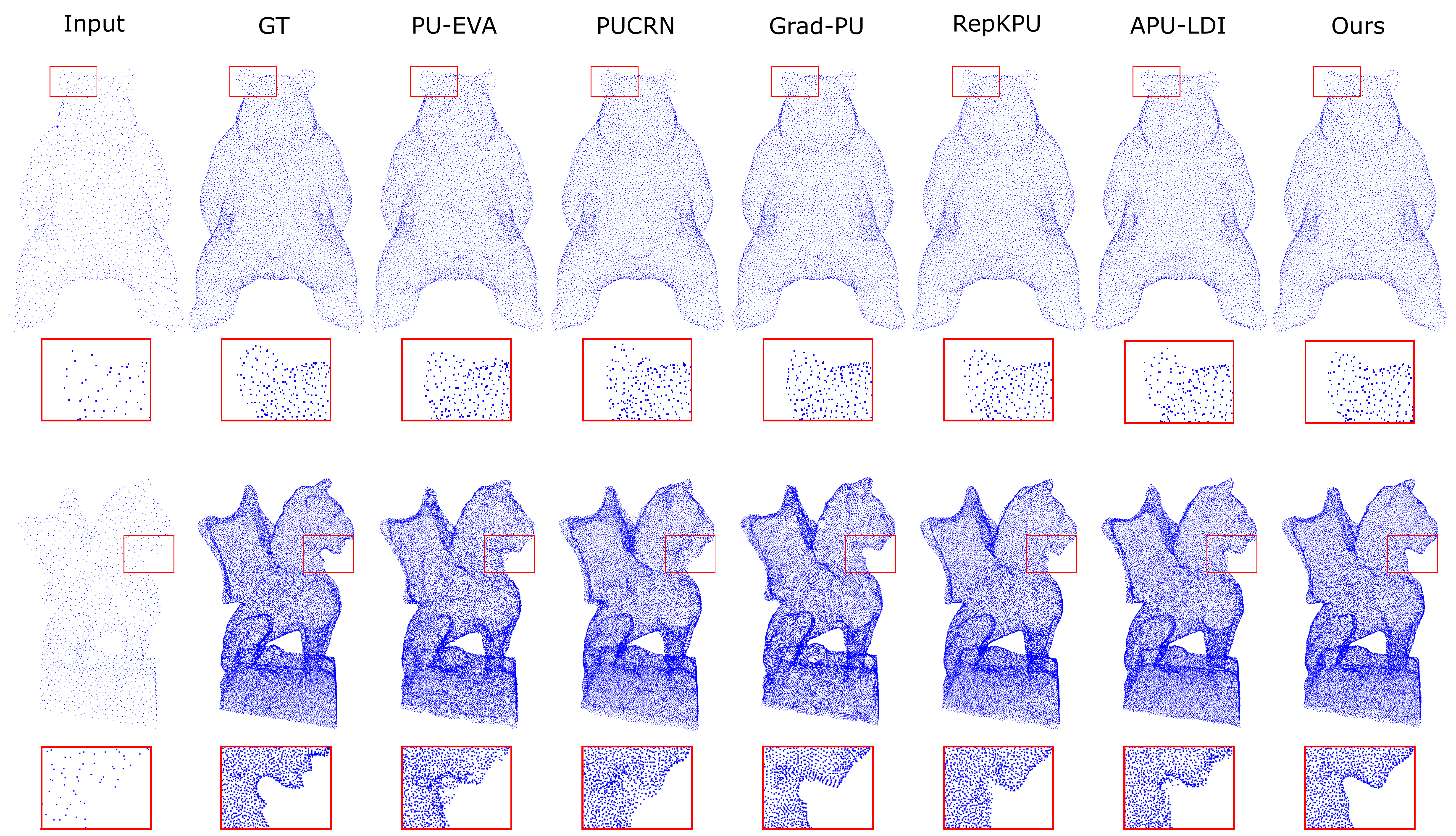}
\vspace{1 mm}
\caption{Qualitative results on the PU-GAN dataset, where the first row is $4\times$ evaluation, the second row is $16\times$ evaluation.
Our results achieve the highest upsampling quality, with fewer outliers, smoother surfaces, and finer-grained details.}
\label{fig:pugan}
\end{figure*}

\section{Experiment}

In this section, we evaluate the performance of our approach in comparison with state-of-the-art point cloud upsampling methods. Experiments are conducted on two synthetic datasets—\textbf{PUGAN}  \cite{li2019pu} and \textbf{PU1K} \cite{qian2021pu}—as well as the real-world \textbf{KITTI} \cite{geiger2013vision} dataset. The results demonstrate the effectiveness of our method in both synthetic and real-world scenarios.

\textbf{Baseline Comparison:} We compare our method with several state-of-the-art approaches, including PU-CRN \cite{du2022point}, PU-EVA \cite{luo2021pu}, Grad-PU \cite{He_2023_CVPR}, RepKPU \cite{rong2024repkpu}, and APU-LDI \cite{li2024APU-LDI}.

\textbf{Implementation Details.}
 All experiments are implemented using PyTorch. We train our
Gaussian network for 100 epochs using the Adam optimizer with an initial learning rate of 0.001 and default settings. An exponential learning rate decay with a factor of 0.7 is applied every 50,000 iterations. Followed by training the full pipeline for 100 epochs with the same settings.

\textbf{Metrics:} We evaluate our method using standard point cloud reconstruction metrics: Chamfer Distance (CD), Hausdorff Distance (HD), and Point-to-Surface Distance (P2F).

\textbf{About PUGAN:} The PUGAN dataset was introduced by \cite{li2019pu} and has since become a benchmark for the point cloud upsampling problem. It comprises 147 3D meshes, with 120 used for training
and 27 for testing. During training, each mesh is segmented into 200 patches, each containing 256 points.
Unlike previous methods that typically train on a 4× upsampling ratio, our approach uses a 6× ratio following the setup in  \cite{luo2021pu}. For testing, we downsample our prediction back to 4x using farthest point sampling. We evaulate our models both on 4× evaluation and 16× evaluation following the setup in  \cite{He_2023_CVPR, rong2024repkpu}. For the 16× evaluation, we upsample the input by 4× and then apply a second 4× upsampling stage. For both evaulations, the number of input points is 2048 pts.

\textbf{About PU1K:} The PU1K dataset was introduced by \cite{qian2021pu}. Compared to PUGAN, PU1K is a larger and more challenging benchmark. It contains 1,147 meshes, with 1,020 used for training and  127 for testing. Of these, 147 meshes are sourced directly from the PUGAN dataset, while
the remaining 1,000 are sampled from the ShapeNet \cite{chang2015shapenet} dataset by selecting 200 shapes from 50 different categories.
For training, alongside the patches from PUGAN, each mesh in PU1K is segmented into 50 patches, resulting in a total of 69,000 training patches. We follow the same procedure as in our PUGAN setup, training with a 6× upsampling ratio and performing 4× evaluation
by downsampling predictions using farthest point sampling.
For testing, we evaulate our method on 4x upsampling ratio following the same procedure in \cite{He_2023_CVPR, rong2024repkpu}.

\textbf{Results on PUGAN:} Our method achieves state-of-the-art performance in the 4$\times$ upsampling evaluation, outperforming all baseline methods in Chamfer distance and ranking second in the remaining metrics, as shown in Table~\ref{tab:pu_gan_compact}. The performance gap is more pronounced in the 16$\times$ evaluation, where our method outperforms all compared methods in both Chamfer and Hausdorff distances. Our qualitative results for the 16 $\times$ evaluation contain fewer outliers than any other approach, highlighting the accuracy of the Gaussian representation as shown in Figure ~\ref{fig:pugan}. 

\begin{table}[t]
\centering
\renewcommand\arraystretch{1.}
\footnotesize
\begin{tabularx}{\linewidth}{lXXX|XXX}
\toprule
\multicolumn{1}{c}{\multirow{2}{*}{\textbf{Method}}} & \multicolumn{3}{c|}{$4\times$ (r=4)} & \multicolumn{3}{c}{$16\times$ (r=16)} \\
\cmidrule(lr){2-4} \cmidrule(lr){5-7}
& \textbf{CD} & \textbf{HD} & \textbf{P2F} & \textbf{CD} & \textbf{HD} & \textbf{P2F}  \\
\midrule
PU-Net \cite{yu2018pu}      & 0.507 & 4.312 & 4.694 & 0.596 & 6.929 & 6.014 \\
PU-GAN  \cite{li2019pu}      & 0.284 & 3.301 & 2.660 & 0.202 & 5.045 & 2.996 \\
Dis-PU  \cite{li2021point}      & 0.278 & 3.447 & 2.343 & 0.180 & 4.888 & 2.560 \\
PU-GCN   \cite{qian2021pu}     & 0.274 & 2.974 & 2.535 & 0.144 & 3.713 & 2.731 \\
APU-SMG \cite{dell2022arbitrary} & 0.296 & 2.404 & 2.492 & 0.155 & 2.749 & 2.573 \\
RepKPU  \cite{rong2024repkpu}     & 0.248 & 2.880 & \cellcolor{red!20}1.906 & 0.107 & 3.345 & 2.068 \\
Grad-PU  \cite{He_2023_CVPR}     & 0.260 & 2.462 & 1.949 & 0.132 & 2.421 & 2.190 \\
PU-VoxelNet \cite{du2024arbitrary}  & \cellcolor{red!20}0.233 & \cellcolor{red!20}1.751 & 2.137 & \cellcolor{yellow!20}\underline{0.091} & 1.726 & 2.301 \\
APU-LDI \cite{li2024APU-LDI}      & \cellcolor{yellow!20}\underline{0.232} & \cellcolor{green!30}\textbf{1.679} & \cellcolor{green!30}\textbf{1.338} & \cellcolor{red!20}0.092 & \cellcolor{yellow!20}\underline{1.504} & \cellcolor{green!30}\textbf{1.544} \\
\midrule
{Ours} 
              & \cellcolor{green!30}{\textbf{0.228}} 
              & \cellcolor{yellow!20}{\underline{1.710}} 
              & \cellcolor{yellow!20}{\underline{1.660}} 
              & \cellcolor{green!30}{\textbf{0.079}} 
              & \cellcolor{green!30}{\textbf{1.443}} 
              & \cellcolor{yellow!30}{\underline{1.720}} \\
\bottomrule
\end{tabularx}
\caption{\textbf{Performance comparison on PU-GAN.} We report CD, HD, and P2F (all $\times 10^{-3}$). Ours achieves strong results across upsampling factors.}
\label{tab:pu_gan_compact}
\end{table}

\begin{figure*}[t]
\centering
\includegraphics[width=0.9\textwidth,keepaspectratio]{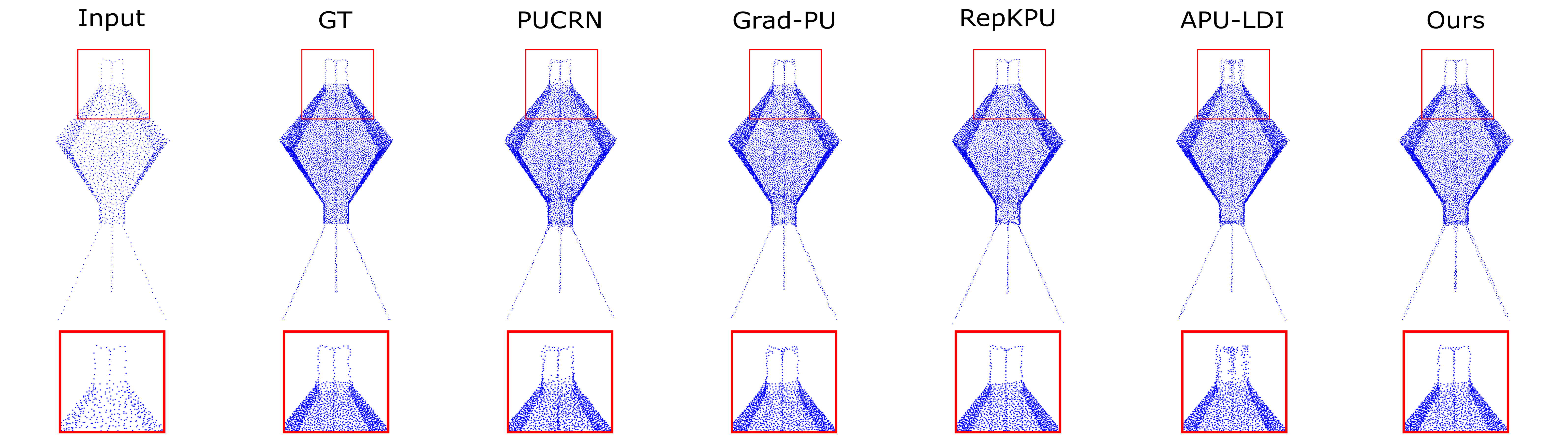}
\vspace{1mm}
\caption{Qualitative results on the PU1K dataset, on $4\times$ evaluation. Our results clearly achieve the highest upsampling quality, with fewer outliers, smoother surface,s and finer-grained details.}
\label{fig:pu1k}
\end{figure*}


\textbf{Results on PU1K:} Our method achieves state-of-the-art performance in the 4× upsampling evaluation, outperforming all baseline methods in Chamfer distance and Hausdorff distance
while ranking third in the P2F, as shown in Table ~\ref{tab:pu1k_final}. Our qualitative results for the 4×
evaluation contain fewer outliers than any other approach and better preserve the underlying shape of object. highlighting the accuracy of the Gaussian representation and the quality
of the output of our network as shown in Figure ~\ref{fig:pu1k}. 

\begin{table}[t]
\centering
\renewcommand\arraystretch{1.1}
\footnotesize
\begin{tabularx}{\linewidth}{lXXX}
\toprule
\textbf{Method} & \textbf{CD} & \textbf{HD} & \textbf{P2F} \\
\midrule
PU-Net \cite{yu2018pu}       & 1.157 & 15.297 & 4.924 \\
MPU \cite{yifan2019patch}    & 0.861 & 11.799 & 3.181 \\
PU-GAN \cite{li2019pu}       & 0.661 & 9.238 & 2.892 \\
Dis-PU \cite{li2021point}    & 0.731 & 9.505 & 2.719 \\
PU-GCN \cite{qian2021pu}     & 0.585 & 7.577 & 2.499 \\
Grad-PU \cite{He_2023_CVPR}  & 0.404 & 3.732 & 1.474 \\
APU-LDI \cite{li2024APU-LDI}               & 0.371 & 3.197 & \cellcolor{yellow!20} \underline{1.111} \\
PU-VoxelNet \cite{du2024arbitrary} 
                             & \cellcolor{red!20}0.338 & \cellcolor{red!20}2.694 & 1.183 \\
RepKPU \cite{rong2024repkpu} 
                             & \cellcolor{yellow!20}\underline{0.327} & \cellcolor{yellow!20}\underline{2.680} & \cellcolor{green!30}\textbf{0.938} \\
\midrule
{Ours}  & \cellcolor{green!30}{\textbf{0.323}} & \cellcolor{green!30}{\textbf{2.593}} & \cellcolor{red!20}{1.176} \\
\bottomrule
\end{tabularx}

\caption{\textbf{Comparison with state-of-the-art methods on PU1K.} Metrics reported: CD, HD, and P2F. Our method achieves competitive performance.}
\label{tab:pu1k_final}
\end{table}

\textbf{Robustness Testing:} To demonstrate the robustness of our approach under such challenging conditions, we conduct two robustness tests using the PUGAN dataset: noise robustenss test, and sparse input test.

\textbf{Noise Robustness Tests:} we evaluate the model’s performance under additive Gaussian noise with standard deviations of 0.01 and 0.02. Noise is added to the input point clouds (2048 points), and results are compared against state-of-the-art methods.
As shown in Table ~\ref{tab:noise_reformat} our model consistently ranks second, only behind APU-LDI \cite{li2024APU-LDI}.
However, unlike APU-LDI—which requires separate training for each test sample, leading to significantly longer runtimes—our method is lightweight and requires training only once. Although the performance gap widens slightly at higher noise levels (e.g., 0.02), this trend is consistent with other geometry-aware methods such as RepKPU \cite{rong2024repkpu}, which also degrade under strong noise due to their reliance on local patch distribution. This behavior is expected, as noisy inputs produce unreliable local features. Nevertheless, our method outperforms all other single-pass upsampling approaches under the same conditions.

\begin{table}[t]
\centering
\renewcommand\arraystretch{1.1}
\footnotesize
\begin{tabularx}{\linewidth}{lXXX|XXX}
\toprule
\multirow{2}{*}{\textbf{Method}} & \multicolumn{3}{c|}{$\tau=0.01$} & \multicolumn{3}{c}{$\tau=0.02$} \\
\cmidrule(lr){2-4} \cmidrule(lr){5-7}
& \textbf{CD} & \textbf{HD} & \textbf{P2F} & \textbf{CD} & \textbf{HD} & \textbf{P2F} \\
\midrule
PU-Net \cite{yu2018pu}       & 0.628 & 8.068 & 9.816 & 1.078 & 10.867 & 16.401 \\
PU-GAN \cite{li2019pu}       & 0.464 & 6.070 & 7.498 & 0.887 & 10.602 & 15.088 \\
Dis-PU \cite{li2021point}    & 0.419 & 5.413 & 6.723 & 0.818 & 9.345 & 14.376 \\
PU-EVA \cite{luo2021pu}      & 0.459 & 5.377 & 7.189 & 0.839 & 9.325 & 14.652 \\
PU-GCN \cite{qian2021pu}     & 0.448 & 5.586 & 6.989 & 0.816 & 8.604 & 13.798 \\
NePs \cite{feng2022neural}   & 0.425 & 5.438 & 6.546 & 0.798 & 9.102 & 12.088 \\
RepKPU \cite{rong2024repkpu} & 0.405 & 5.013 & 6.890 & 0.819 & 9.546 & 14.279 \\
Grad-PU \cite{He_2023_CVPR}  & 0.414 & 4.145 & 6.400 & 0.766 & 7.339 & 11.534 \\
APU-LDI \cite{li2024APU-LDI} & \cellcolor{green!30}\textbf{0.339} & \cellcolor{green!30}\textbf{3.089} & \cellcolor{green!30}\textbf{5.167} & \cellcolor{green!30}\textbf{0.622} & \cellcolor{green!30}\textbf{6.485} & \cellcolor{green!30}\textbf{10.984} \\
\midrule
Ours  & \cellcolor{yellow!20}{\underline{0.353}} & \cellcolor{yellow!20}{\underline{3.196}} &\cellcolor{yellow!20}{\underline{5.932}} & \cellcolor{yellow!20}{\underline{0.742}} & \cellcolor{yellow!20}{\underline{6.840}} & \cellcolor{yellow!20}{\underline{11.757}} \\
\bottomrule
\end{tabularx}
\caption{\textbf{Robustness to noise on PU-GAN.} All metrics are scaled by $10^{-3}$. Our method remains competitive under noise perturbation.}
\label{tab:noise_reformat}
\end{table}

\begin{figure}[h]
\centering
\includegraphics[width=.75\columnwidth,keepaspectratio]{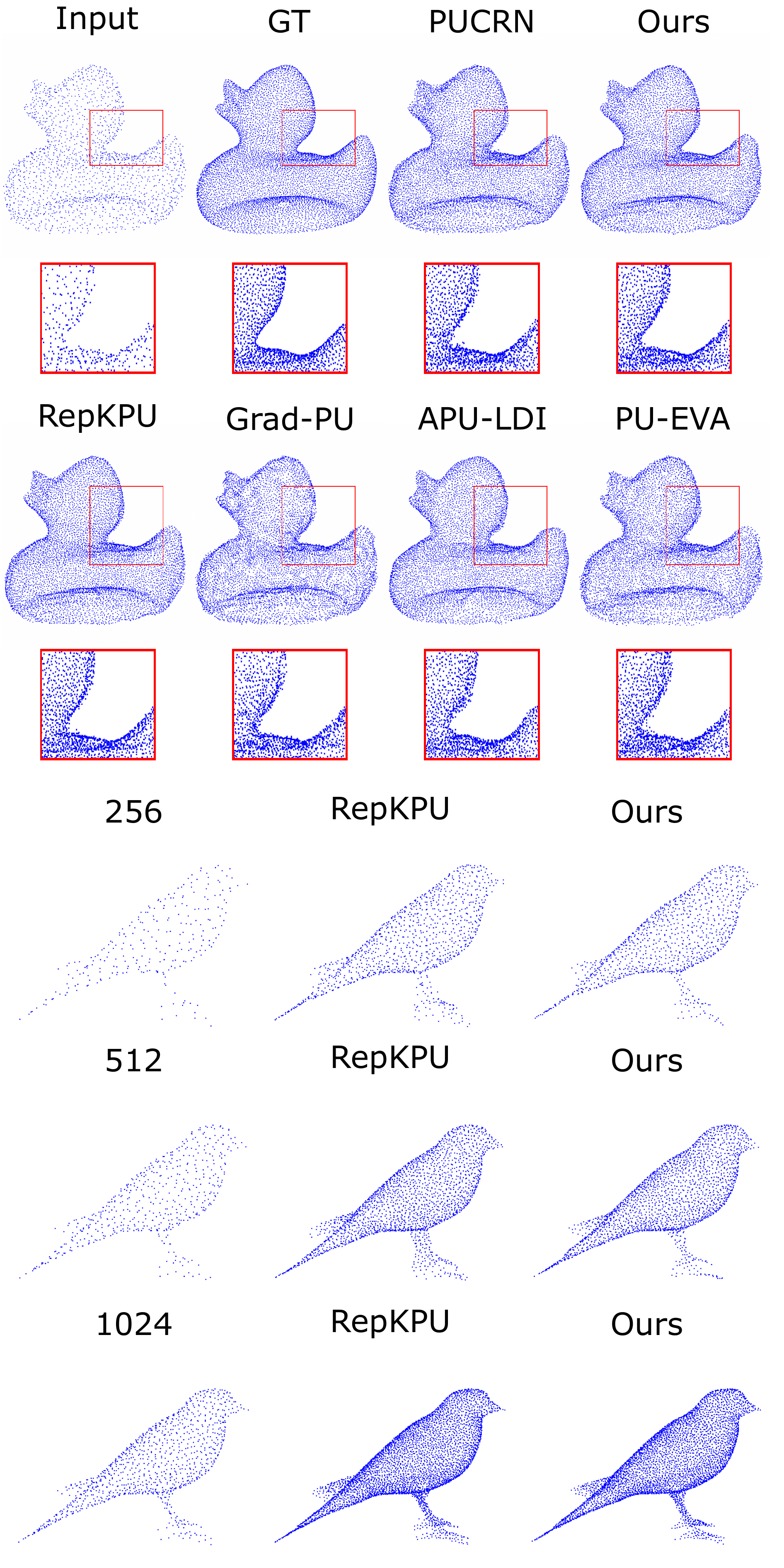}
\vspace{1mm}
\caption{Qualitative results from the robustness test on the PUGAN dataset under $4\times$ evaluation. Top, the results for the $4\times$ evaluation with added noise level $\tau$ = 0.01. Bottom, we compare various inputs against REPKPU. Despite the highly sparse input, our method yields a more well-defined surface.}
\label{fig:robustness}
\end{figure}

\textbf{Sparsity Robustness Test:} we evaluate performance across varying input sparsity levels, using inputs of 256, 512, and 1024 points. We test at both 4× and 16× upsampling rates and report the average results over all testing meshes as shown in Table ~\ref{tab:sparse_inputs}. Our model consistently outperforms existing methods at every sparsity level. In particular, the qualitative results demonstrate our method’s ability to preserve structural details even at extremely sparse input, highlighting its practical applicability as shown in Figure \ref{fig:robustness}.

\begin{table}[H]
\centering
\renewcommand\arraystretch{1.1}
\footnotesize
\begin{tabularx}{\linewidth}{lXXX}
\toprule
\textbf{Method} & \textbf{CD} & \textbf{HD} & \textbf{P2F} \\
\midrule
RepKPU \cite{rong2024repkpu}     & 0.9150 & 11.63 & 5.95 \\
PUCRN \cite{du2022point}         & 1.2046 & 18.99 & 8.47 \\
PUEVA \cite{luo2021pu}           & 1.3369 & 17.59 & 7.87 \\
Grad-PU \cite{He_2023_CVPR}      & 1.6907 & 14.32 & 9.98 \\
\midrule
{\textbf{Ours}} 
                                 & \cellcolor{green!30}\textbf{0.8803} 
                                 & \cellcolor{green!30}\textbf{7.50} 
                                 & \cellcolor{green!30}\textbf{4.96} \\
\bottomrule
\end{tabularx}
\caption{\textbf{Comparison on sparse inputs} (256, 512, 1024 points) under $4\times$ and $16\times$ upsampling. Our method shows strong generalization on low-resolution inputs.}
\label{tab:sparse_inputs}
\end{table}

\begin{figure*}[t!]
\centering
\includegraphics[width=\textwidth]{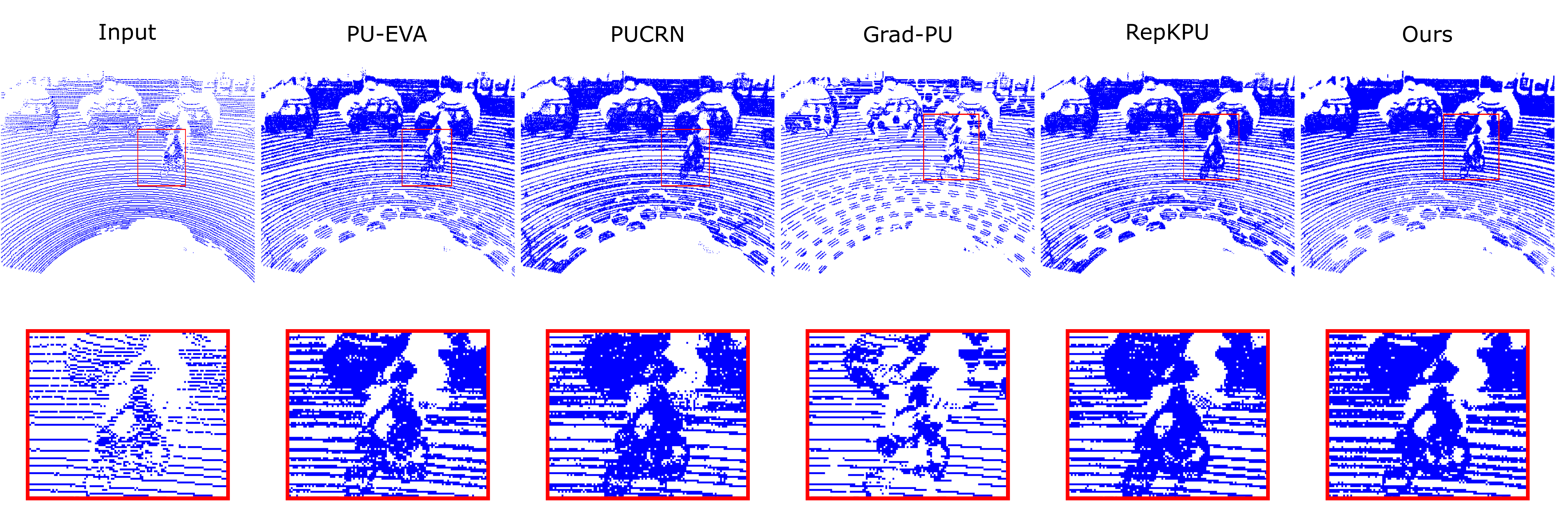}
\caption{Qualitative results on the KITTI dataset, on $4\times$ evaluation. Our results clearly achieve the highest output quality with fewer outliers. }
\label{fig:kitti}
\end{figure*}

\begin{figure*}[t!]
\centering
\includegraphics[width=\textwidth]{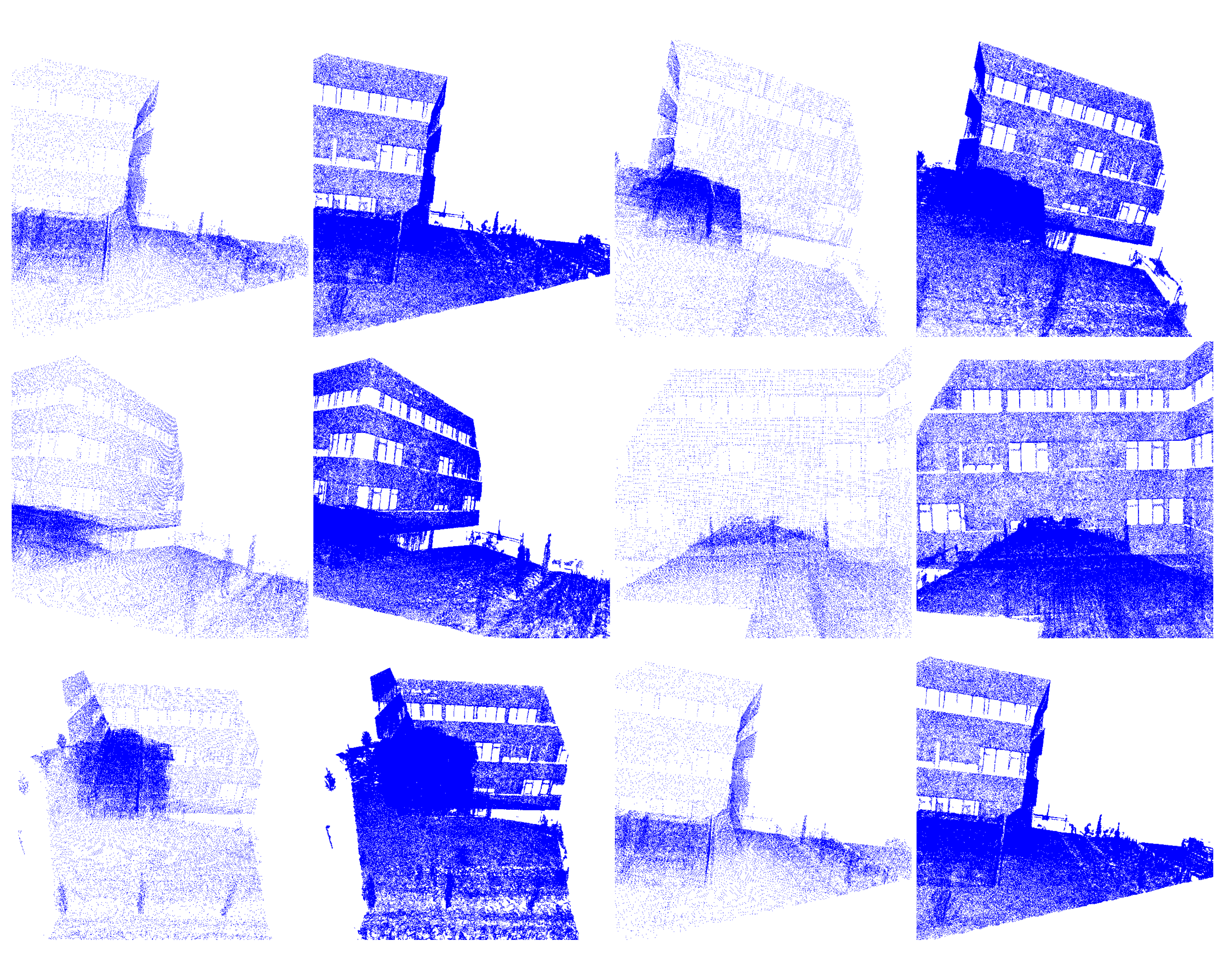}
\caption{Qualitative results on our internal scans. Our results preserve the
shape of the scan and contains few outliers. }
\label{fig:frany}
\end{figure*}


\textbf{Real-World Scenario}:
We evaluate our method on the KITTI dataset shown in Figure \ref{fig:kitti}, comparing results with baselines to demonstrate real-world applicability. Since there is no ground truth dense point clouds in KITTI, we only provide qualitative results. Our qualitative results show a superior performance with fewer outliers and better shape preservation. We also present our upsampling results on internal scans from Fraunhofer IPM, as illustrated in Figure~\ref{fig:frany}. Our method produces visually appealing results while preserving the geometry of the buildings and the surrounding area. This demonstrates PU-Gaussian’s consistent geometric fidelity and robustness to real-scene variability.


\textbf{Conclusion:}
PU-Gaussian method demonstrates strong performance under a variety of conditions, consistently outperforming state-of-the-art methods across several quantitative metrics. It also produces visually coherent and high-quality results on both synthetic and real-world data. We have developed a lightweight and efficient module that achieves competitive results. The proposed Gaussian representation is especially effective compared to alternative feature expansion techniques such as Folding and Node Shuffling, as evidenced by our performance improvements over PU-GCN \cite{qian2021pu} and other recent methods. However, PU-Gaussian has two primary limitations. First, its sampling-based nature introduces stochasticity in the output. Our refinement network, however, helps reduce this variability. We propose that future work should explore a learnable Gaussian-to-point module capable of generating deterministic outputs directly, thereby eliminating the need for post-refinement. Second, like other geometry-based upsampling methods, the performance of PU-Gaussian is degraded in the presence of high noise due to a decrease in the local feature quality. Addressing this challenge remains an open problem for future research.


\section*{Acknowledgments}
This work was supported as a Fraunhofer FLAGSHIP PROJECT. 
\newpage
\clearpage

{
    \small
    \bibliographystyle{ieeenat_fullname}
    \bibliography{main}
}


\end{document}